%% file: edm_paper.tex
\documentclass{edm_article}

\usepackage{capt-of}
\usepackage{listings}
\usepackage{xcolor}
\usepackage{amsmath}
\usepackage{amsfonts}
\usepackage{booktabs, multirow}
\usepackage{colortbl}
\usepackage{hyperref}
\usepackage{url}
\usepackage{wrapfig}

\begin{document}

\title{Interpreting Latent Student Knowledge Representations in Programming Assignments}

\numberofauthors{2}
\author{
\alignauthor
    Nigel Fernandez, Andrew Lan\\
    \affaddr{University of Massachusetts Amherst}\\
    \email{\{nigel,andrewlan\}@cs.umass.edu}
}

\maketitle

\begin{abstract}
Recent advances in artificial intelligence for education leverage generative large language models, including using them to predict open-ended student responses rather than their correctness only. However, the black-box nature of these models limits the interpretability of the learned student knowledge representations. In this paper, we conduct a first exploration into interpreting latent student knowledge representations by presenting InfoOIRT, an Information regularized Open-ended Item Response Theory model, which encourages the latent student knowledge states to be interpretable while being able to generate student-written code for open-ended programming questions. InfoOIRT maximizes the mutual information between a fixed subset of latent knowledge states enforced with simple prior distributions and generated student code, which encourages the model to learn disentangled representations of salient syntactic and semantic code features including syntactic styles, mastery of programming skills, and code structures. Through experiments on a real-world programming education dataset, we show that InfoOIRT can both accurately generate student code and lead to interpretable student knowledge representations.  
\end{abstract}


%
%
%
%
%
%

\section{Introduction}

Open-ended problems, which require students to produce free-form responses, either as short answers or essays~\cite{attali2006automated} or even code~\cite{shi2022code}, serve as a highly meaningful form of assessment and complements closed-form problems such as multiple-choice questions~\cite{ozuru2013comparing}. These questions often require students to detail their reasoning process and offer educators a deeper look into their knowledge states. Past work has shown that students' open-ended responses to such questions contain useful information on their knowledge states, e.g., having misconceptions~\cite{brown,feldman,smith} or generally lacking sufficient knowledge~\cite{anderson}. Until recently, however, research has mostly focused on the automated \emph{scoring} of open-ended responses, either via classification methods~\cite{nigel-scoring,taghipour2016neural,zhang2022automatic} or clustering~\cite{parrdosembed} and providing corresponding feedback~\cite{hanawa2021exploring,heickal2024generating,kumar2024improving,hunter-math,roscoe2013developing,scarlatos2024improving}. However, relatively little has been done towards developing student response models that estimate their knowledge from open-ended responses; existing models such as item response theory (IRT)~\cite{irtbook}, knowledge tracing~\cite{kt}, and factor analysis~\cite{pfa} primarily analyze close-ended responses or graded ones, which are either binary-valued or nominal/ordinal. These models are fundamentally limited for open-ended problems since they cannot fully extract detailed information on student knowledge contained in their free-form responses. See Section~\ref{sec:related_work} for a detailed discussion on related work.

Recent advances in pre-trained generative large language models (LLMs)~\cite{bubeck2023sparks} provide an opportunity to gain deeper insights into student knowledge by analyzing their free-form responses. Most existing works use text embedding models to summarize open-ended responses only as input into knowledge tracing models~\cite{su2018exercise}, not fully utilizing the generative capabilities of LLMs. The only recent work that combines generative LLMs with an underlying student response model is open-ended knowledge tracing (OKT)~\cite{okt}, which uses a knowledge tracing model to track the change in student knowledge state over time, and then injects that knowledge state together with the textual problem statement as input to a generative LLM to predict a student's open-ended response as output. Applied to student-written code to programming problems, OKT shows that learned underlying latent student knowledge states have some correlation with student-written code. 
Despite some early promise, a key limitation of OKT is the interpretability of the latent student knowledge space; there is no clear way to isolate certain elements in these vectors that capture key aspects of student code: ones that reflect their knowledge of key programming knowledge concepts, ones that reflect certain bugs/misconceptions, or even ones that capture distinct coding styles. Since there is no prior enforced on the structure of the latent knowledge space, the black-box nature of LLMs would likely lead to entangled representations that are highly predictive of student responses but hard to interpret. 

%
%

\subsection{Contributions}
In this paper, we present a first attempt at interpreting latent student knowledge states in models of open-ended responses, specifically on code that students write for open-ended programming questions. Our contributions are:
\begin{enumerate}
    \item We develop InfoOIRT~\footnote{Code: \url{https://github.com/umass-ml4ed/InfoOIRT}}, an Information-regularized Open-ended IRT model, which predicts student-written code for open-ended programming questions with a focus on learning interpretable student knowledge representations. Inspired by InfoGAN~\cite{chen2016infogan}, InfoOIRT \emph{maximizes mutual information between a fixed subset of latent knowledge states enforced with simple prior distributions and generated student code} to encourage the latent factors to learn disentangled representations of salient code features. Although our idea can potentially be applied to other subjects with open-ended questions, we ground our analysis in Computer Science (CS) education. 
    \item We conduct quantitative experiments on a real-world student code dataset to show that information regularization does not impact the ability of InfoOIRT to accurately predict student code compared to baselines.
    \item We conduct qualitative analyses to interpret the learned student knowledge representations. Using a combination of both continuous and discrete latent student knowledge state factors, we present examples of generated student code highlighting the salient syntactic and semantic features captured by these states. 
\end{enumerate}

%
%
%
%
%
%

\section{Related Work}
\label{sec:related_work}
%
%

\subsection{Interpretable Representation Learning}
There exists prior work in learning interpretable representations for the underlying processes of image~\cite{image_interpretable_2,image_interpretable_1} and text~\cite{text_interpretable} generation. A seminal method in unsupervised representation learning, InfoGAN~\cite{chen2016infogan} aims to learn disentangled representations, one which explicitly represents the salient features of the data as easily interpretable factors (e.g., number, orientation, and stroke thickness in hand-written digits), using an information-based regularization in the training objective. InfoOIRT extends this idea to learn interpretable representations in LLMs for code generation, specifically student responses to programming problems.

%
%

\subsection{Student Modeling}

There exist many models of student knowledge, depending on how they characterize both latent knowledge states and observed responses. For latent knowledge states, the highly interpretable Bayesian knowledge tracing model treats them as binary-valued, i.e., whether a student masters a skill or not. 
Factor analysis-based methods~\cite{das3h,pfa} use a set of hand-crafted features to summarize past student activities and represent student knowledge, before relying on IRT models to predict student responses from these features. 
On the contrary, deep learning-based KT methods~\cite{long2021tracing,rkt,dkt,saint+,dkvmn} treat student knowledge as latent vectors in deep neural networks, resulting in models that excel at future performance prediction but have limited interpretability.

For observed responses, despite most existing models treating them as binary-valued, i.e., correct/incorrect, there exist some models that analyze the exact student response including multiple-choice options~\cite{ot} and partial credits~\cite{neilkt}. 
In general, one can use polytomous IRT models~\cite{polytomous} as the response prediction component in KT methods to predict categorical-valued (such as options in multiple-choice questions) and ordinal-valued (such as partial credit) responses~\cite{sparfatag}. In the programming domain,~\cite{ncsupkt,pkt} use code embedding techniques to convert student-written code into vectors to help student models track their progress. However, they do not use generative LLMs to predict student code. 

%
%

\subsection{Program Synthesis and CS Education}

There exist many works applying program synthesis techniques for computer science education to generate (possibly buggy) student code~\cite{okt,mpi_synth}, generate new problems~\cite{task_synth} with code explanations~\cite{sarsa2022automatic}, generate student-code guided test cases~\cite{kumar2024using}, provide real-time hints~\cite{hints}, and suggest bug fixes~\cite{koutcheme2023automated}. However, the black-box nature of these models provides limited interpretability.

%
%
%
%
%
%

\section{Interpretable Open-ended IRT}

%
%

\subsection{Problem Formulation and OIRT}

Item response theory (IRT)~\cite{irt} involves diagnosing a student's mastery of knowledge components/skills/concepts from their responses to problems, where we assume a student's knowledge state is static, i.e., it does not change as they respond to problems. For open-ended item response theory (OIRT), we need two essential components. First, a knowledge estimation (KE) component that estimates a student $j$'s knowledge state from the set of student code submissions $c_{ij}$ to problems $p_i$ denoted by $\{(p_i, c_{ij})\}$, i.e., $h_j = \text{KE}\left( \{(p_i, c_{ij})\}\right)$. Second and more importantly, a response generation (RG) component that predicts student $j$'s open-ended code submission to a target problem $p_k$ using a generative model, i.e., $c_{kj} = \text{RG}\left(p_k, h_j\right)$. This generation model is the key difference between OIRT and traditional IRT: our goal is to predict the code a student would write for an open-ended programming problem via a generative model, rather than simply predicting its correctness. 

We denote the student's latent knowledge as a $d$-dimensional vector $h_j$ for every student $j$. This setup is similar to learning a multidimensional student ability parameter in IRT. OIRT leverages generative language models and employs a text-to-code finetuned GPT-2~\cite{gpt2} model. A problem $p_k$ is tokenized by GPT-2 into a sequence of $M$ tokens where each token has a $768$-dimensional embedding, i.e., $\bar{p}_{m} \in \mathcal{R}^{768}$ for $m = 1, \ldots, M$ (here, we drop the problem index $k$). We inject student $j$'s knowledge state $h_j$ by replacing the raw problem token embeddings with knowledge-guided embeddings using a linear alignment function $f$, i.e., $p_m = f(\bar{p}_{m}, h_j)$, similar to~\cite{okt}. The predicted student code is generated autoregressively using GPT-2 given the knowledge-guided problem embeddings as input. OIRT jointly learns the student knowledge states and the fine-tuned GPT-2 parameters together with the linear alignment function. The objective for one student code submission $c_{kj}$ by student $j$ with knowledge state $h_j$ to problem $p_k$, consisting of $N$ tokens, is: $\mathcal{L}_{\text{OIRT}} = \textstyle\sum_{n=1}^{N} -\log P_{\theta}(c_{kj}^n | p_k, h_j, \{c_{kj}^{n'}\}_{n'=1}^{n-1})$, where $\theta$ denotes the learnable parameters of the KE and RG components. The final objective is the sum of this loss $\mathcal{L}_{\text{OIRT}}$ over code submissions by all students to all problems.

%
%
%
%
%
%

\subsection{InfoOIRT: Information-regularized OIRT}

\begin{figure}
\Description{Image of the Architecture of Information-regularized Open-ended IRT (InfoOIRT) Model.}
\centering
\includegraphics[width=\linewidth]{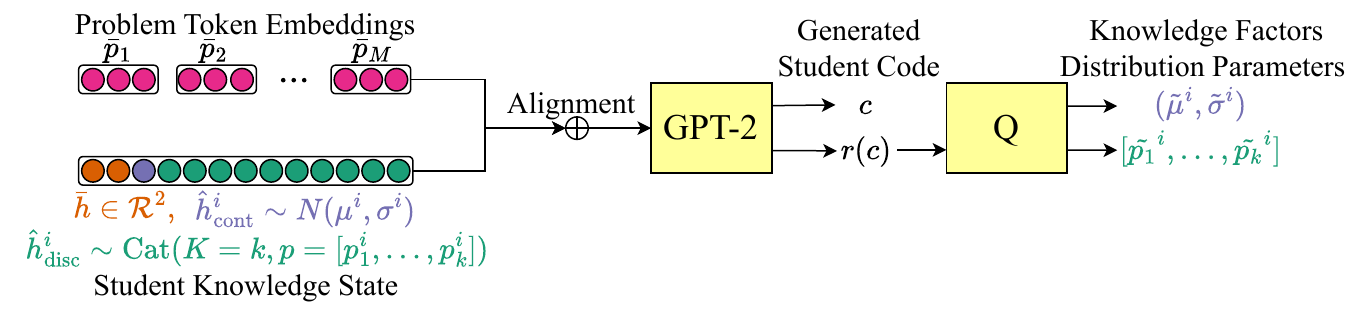}
\caption{InfoOIRT Model Architecture}
\label{fig:infooirt}
\end{figure}

One key limitation of OIRT is that the learned student knowledge states are hard to interpret and associate with different programming skills. We present a simple modification of OIRT, inspired by InfoGAN~\cite{chen2016infogan}, to learn interpretable and meaningful latent student knowledge states. The idea is to \emph{maximize the mutual information between a fixed subset of the student knowledge state dimensions enforced with simple prior distributions and the generated student code}. These dimensions help us discover semantic and meaningful hidden representations of student code. 

We now detail our InfoOIRT model visualized in Figure~\ref{fig:infooirt}. 
Following InfoGAN~\cite{chen2016infogan}, we decompose a latent student knowledge state $h$ into two parts: 1) $\bar{h}$, which represents the incompressible student knowledge state, and 2) $\hat{h}$, which consists of simple and interpretable latent factors $\hat{h}^1, \ldots, \hat{h}^K$, to represent the salient structured semantic features of student written code. However, GPT-2 could \emph{ignore} these additional latent factors $\hat{h}$ and simply generate student code $c$ with a probability distribution satisfying $P(c | \hat{h}) = P(c)$. 
We, therefore, impose an \emph{information-theoretic regularization} to encourage high mutual information between latent factors $\hat{h}$ and the GPT-2 generator distribution $G(\bar{h}, \hat{h}, p)$ which generates student code $c$ corresponding to a student with knowledge state $h$. This regularization encourages the latent factors to explicitly contain information that dictates the variation in student code, disentangling the interpretable dimensions from incompressible noise. Specifically, mutual information is defined as $I(\hat{h}, G(\bar{h}, \hat{h}, p)) = H(\hat{h}) - H(\hat{h} | G(\bar{h}, \hat{h}, p))$. 
Intuitively, mutual information is maximized when the uncertainty, i.e., entropy, in $\hat{h}$ given $c$ is minimized, meaning that the \emph{information in the simple latent knowledge factors $\hat{h}$ should not be lost in the generation of the student code $c$}, thereby reducing the likelihood of GPT-2 ignoring the latent factors during student code generation. 
However, maximizing the mutual information directly is hard since it requires access to the posterior $P(\hat{h} | c)$. 
Therefore, we use a variational lower bound of mutual information following~\cite{chen2016infogan}: $I(\hat{h}, G(\bar{h}, \hat{h}, p)) \geq L_{I}(G, Q) = \mathbb{E}_{\hat{h} \sim P(\hat{h}), c \sim G(\bar{h}, \hat{h}, p)}[\log Q(\hat{h} | c)] + H(\hat{h})$.
We treat $H(\hat{h})$ as constant for simplicity and arrive at the regularized InfoOIRT loss for a single student code: $\mathcal{L}_{\text{infoOIRT}} = \mathcal{L}_{\text{OIRT}} - \lambda L_{I}(G, Q)$.
The final objective is the sum of $\mathcal{L}_{\text{infoOIRT}}$ over code submissions by all students to all problems. We refer readers to~\cite{chen2016infogan} for details.

Similar to OIRT, InfoOIRT also uses a static knowledge state vector $\bar{h} \in \mathcal{R}^{d_{\text{bar}}}$ for every student to represent incompressible noise. We chose to model the interpretable latent factors $\hat{h}^i$ as having $d_{\text{cont}}$ dimensions, each having a simple Gaussian distribution, $\hat{h}^i_{\text{cont}} \sim N(\mu^i, \sigma^i)$, and $d_{\text{disc}}$ discrete dimensions, each having a simple categorical distribution with $k$ classes, $\hat{h}^i_{\text{disc}} \sim \text{Cat}(K=k, p=[p_1^i, \ldots, p_k^i])$. We learn the student-specific distribution parameters of each of the latent factors $\hat{h}^i$, namely the mean and standard deviation $(\mu^i, \sigma^i)$ for continuous dimensions, and the categorical distribution parameters $(p_{1}^{i}, \ldots, p_{k}^{i})$ for discrete dimensions. To estimate a student's knowledge state, we sample each latent factor $\hat{h}^i$ using the current learned student-specific distribution parameters. 
We then concatenate the current learned student-specific $\bar{h}$ vector with this sampled $\hat{h}$ vector to get the student's final knowledge state $h$.
We parameterize the auxiliary distribution $Q$ as a fully connected neural network, which takes as input a representation of the generated student code $c$ and outputs the parameters of the distribution $Q(\hat{h} | c)$. We use the mean of the hidden states of the last layer of GPT-2 as a proxy representation $r(c)$ of the generated student code $c$.  
The learnable parameters in InfoOIRT include the Q model, the student-specific incompressible knowledge state $\bar{h}$ and the student-specific distribution parameters for each dimension of the latent factors $\hat{h}$, in addition to those in OIRT. 

%
%
%
%
%
%

\section{Experiments and Results}

%
%

\subsection{Dataset, Metrics, Implementation Details}

We ground our analysis on the CSEDM dataset\footnote{\url{https://sites.google.com/ncsu.edu/csedm-dc-2021}}, a real-world programming education dataset containing $46,825$ student code submissions from $246$ college students on $50$ open-ended Java programming problems collected over a semester. 
Following OKT~\cite{okt}, we quantitatively evaluate generated student code using the popular CodeBLEU~\cite{codebleu} metric, which measures syntactic and semantic similarity between generated and actual student codes. 
We report the average test loss of GPT-2 across generated code tokens.
To test whether InfoOIRT simply memorizes the training data, we measure diversity in the generated student codes using the dist-N metric~\cite{distn}, which computes the ratio of unique $N$-grams in the generated codes over all N-grams, with $N=1,2,3$.
For OIRT training, we follow the setup in~\cite{okt}.
For InfoOIRT, we chose to model the latent factors in student knowledge states $\hat{h}$, with $1$ continuous dimension and $10$ discrete dimensions, each having two classes and thereby encouraged to act as binary switches of programming knowledge mastery states/syntactic styles. 
We learn a student-specific $2$-dimensional static knowledge state $\bar{h}$. For a fair comparison with OIRT, we learn a student-specific $23$-dimensional $\bar{h}$ in OIRT and use the same hyperparameters in both models (see Appendix~\ref{sec:supp_experiments}). 

%
%

\subsection{Quantitative Results}

\begin{table*}[!htb]
\centering
\caption{Experimental results on the CSEDM test set. Our InfoOIRT model is competitive with the baseline OIRT model.} 
\label{tab:results}
\scalebox{.66}{
\begin{tabular}{p{0.19\linewidth}p{0.13\linewidth}p{0.13\linewidth}p{0.13\linewidth}p{0.17\linewidth}p{0.17\linewidth}p{0.13\linewidth}p{0.13\linewidth}p{0.13\linewidth}}

\toprule

\multirow{2}{*}{Model} & \multicolumn{3}{c}{Knowledge State} & \multicolumn{2}{c}{Code Quality} &  \multicolumn{3}{c}{Code Diversity}\\
\cmidrule{2-9}
&$|\bar{h}|$ & $|\hat{h}_{\text{cont}}|$ & $|\hat{h}_{\text{disc}}|$ & CodeBLEU $\uparrow$ & Test Loss $\downarrow$ & Dist-$1$ $\uparrow$ & Dist-$2$ $\uparrow$ & Dist-$3$ $\uparrow$ \\

\midrule

\rowcolor{gray!21} \multicolumn{9}{c}{Main Models}\\

OIRT & $23$ & - & - & $0.597$ & $0.200$ & $0.396$ & $0.712$ & $0.825$ \\
InfoOIRT & $2$ & $1$ & $10$ & $0.601$ & $0.205$ & $0.394$ & $0.712$ & $0.827$\\

\rowcolor{gray!21} \multicolumn{9}{c}{Ablation: Increasing Uninterpretable Knowledge Dimensions $|\bar{h}|$}\\
OIRT & $64$ & - & - & $0.609$ & $0.202$ & $0.399$ & $0.717$ & $0.830$\\
OIRT & $256$ & - & - & $0.607$ & $0.204$ & $0.404$ & $0.719$ & $0.829$\\
InfoOIRT & $64$ & $1$ & $10$ & $0.611$ & $0.199$ & $0.402$ & $0.721$ & $0.832$\\
InfoOIRT & $256$ & $1$ & $10$ & $0.613$ & $0.200$ & $0.400$ & $0.718$ & $0.829$\\

\rowcolor{gray!21} \multicolumn{9}{c}{Ablation: Increasing Interpretable Knowledge Factors $|\hat{h}|$}\\
InfoOIRT & $2$ & $64$ & $64$ & $0.507$ & $0.213$ & $0.383$ & $0.695$ & $0.812$\\
InfoOIRT & $2$ & $256$ & $256$ & $0.510$ & $0.214$ & $0.398$ & $0.710$ & $0.824$\\

\rowcolor{gray!21} \multicolumn{9}{c}{Ablation: InfoOIRT with Continuous Knowledge Factors $|\hat{h}_{\text{cont}}|$ or Discrete Knowledge Factors $|\hat{h}_{\text{disc}}|$ Only}\\
InfoOIRT & $2$ & $1$ & $0$ & $0.606$ & $0.201$ & $0.397$ & $0.717$ & $0.831$\\

InfoOIRT & $2$ & $0$ & $10$ & $0.539$ & $0.212$ & $0.393$ & $0.706$ & $0.822$\\

\bottomrule
\end{tabular}
}
\end{table*}

\begin{figure}
\Description{Graph showing mutual information is maximized with an information-regularized objective.}
\centering
\includegraphics[width=0.55\linewidth]{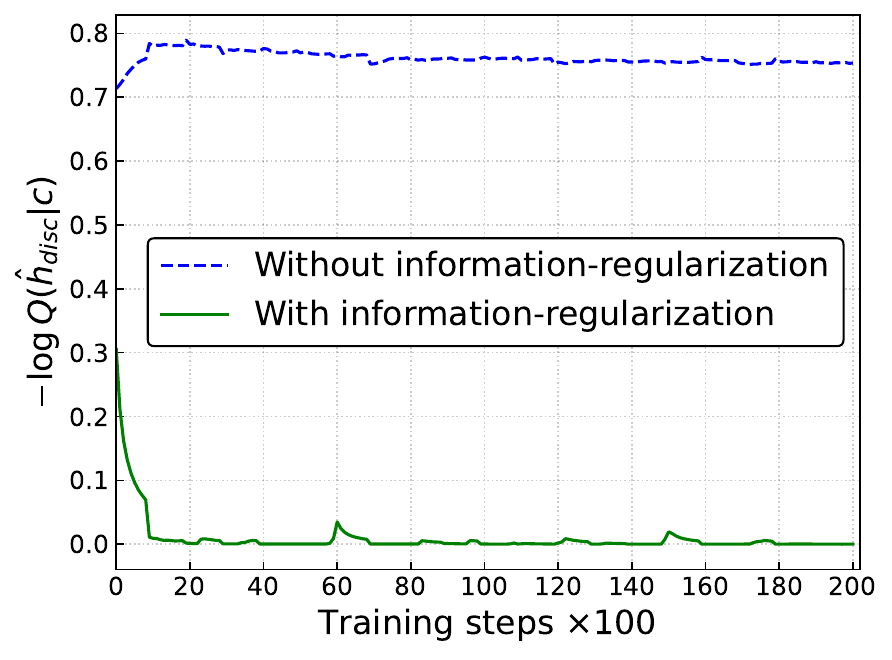}
\caption{Mutual Information Maximization}
\label{fig:loss}
\end{figure}

We show the performance of both OIRT and InfoOIRT on all metrics on the CSEDM test set in Table~\ref{tab:results}. 
We see that InfoOIRT exhibits competitive performance to OIRT. Therefore, the addition of mutual information-based regularization in the objective provides interpretability through simple latent factors $\hat{h}$ without sacrificing code prediction accuracy.
We also show various model ablations in Table~\ref{tab:results}. We observe diminishing performance returns when increasing the number of dimensions in $\bar{h}$. We therefore use a small number of dimensions in $\bar{h}$ in our models to prioritize interpretability without sacrificing performance. We also see performance degradation with a high number of interpretable knowledge factors $\hat{h}$, especially discrete knowledge factors $\hat{h}_{\text{disc}}$. Although $\hat{h}_{\text{disc}}$ provides interpretability, these factors possibly also oversimplify the model by imposing additional constraints, thereby reducing the flexibility of the model. Therefore, our choice of the number of interpretable knowledge factors reflects a balance between performance and interpretability. 
To test whether mutual information is maximized, in Figure~\ref{fig:loss}, we show the negative log-likelihood of our $Q$ model which is quickly minimized with an information-regularized objective indicating high mutual information between latent factors $\hat{h}$ and generated student code $c$. However, an equivalent InfoOIRT model without this regularization objective exhibits low mutual information.

%
%
%
%
%
%

\subsection{Qualitative Results}

\begin{table*}[!htb]
\centering
\caption{Variation in generated code for variation in $\hat{h}_{\text{disc}}$ and $\hat{h}_{\text{cont}}$.}
\label{tab:qualitative_results}
\scalebox{.75}{
\begin{tabular}{ p{0.63\linewidth} | p{0.63\linewidth} }
\toprule
\rowcolor{gray!21} \multicolumn{2}{c}{
Switching $\hat{h}_{\text{disc}}^9$ from $0$ to $1$ results in change in if-else nesting on a subset of problems attempted by one student.}\\
\midrule
$\hat{h}_{\text{disc}}^9 = 0$ Generates Code With If-Else Nesting & $\hat{h}_{\text{disc}}^9 = 1$ Generates Code Without If-Else Nesting\\
\midrule
\input{student_codes/conciseness_11} & \input{student_codes/conciseness_12} \\
\midrule
\input{student_codes/conciseness_41} & \input{student_codes/conciseness_42} \\

\midrule
\rowcolor{gray!21} \multicolumn{2}{c}{
Switching $\hat{h}_{\text{disc}}^6$ from $0$ to $1$ results in change in indentation style on a subset of problems attempted by one student.}\\
\midrule
$\hat{h}_{\text{disc}}^6 = 0$ Generates Code with ``K\&R'' Indentation Style & $\hat{h}_{\text{disc}}^6 = 1$ Generates Code with ``Allman'' Indentation Style \\
\midrule
\input{student_codes/style_11} & \input{student_codes/style_12} \\
\midrule
\input{student_codes/style_41} & \input{student_codes/style_42} \\

\midrule
\rowcolor{gray!21} \multicolumn{2}{c}{
Extreme variation in $\hat{h}_{\text{cont}}$ results in code with conditionals for easier problems shifting to code with loops for harder problems, and vice versa.}\\
\midrule
Student Code for Problems with Conditional Constructs & Student Code for Problems with Looping Constructs \\
\midrule
\input{student_codes/problem_type_11} & \input{student_codes/problem_type_12} \\

\bottomrule
\end{tabular}
}
\end{table*}

%
%

\subsubsection{Discrete Latent Knowledge Factors}
\label{sec:discrete_latent_factors}

We manipulate the learned discrete latent factors $\hat{h}_{\text{disc}}$, each from a binary categorical distribution. For each of the ten factors, we vary the binary class of that factor, keeping the remaining factors constant and set to their learned class, and analyze the resulting variation in generated code. For some discrete factors, these changes reflect different styles (indentation, spacing between function arguments), mastery of programming skills (correct or incorrect codes with different bugs), or code structure (for loops to while loops, if-else with nesting to without nesting). For example, as shown in Table~\ref{tab:qualitative_results}, for one student, switching $\hat{h}_{\text{disc}}^9$ from class $0$ to $1$, resulted in change from code using if-else with nesting to code using if-else without nesting, while for another student, switching $\hat{h}_{\text{disc}}^6$ from class $0$ to $1$ resulted in change in indentation style. We note that such changes are not found in all students and all problems: problems in the CSEDM dataset cover a wide range of programming skills and many of these changes apply to few problems. 
For easier problems with shorter student codes, we observe minimal variation since InfoOIRT is often able to predict the exact student code written (which is correct). In these cases, InfoOIRT prioritizes the $\mathcal{L}_{\text{OIRT}}$ loss for generation performance and possibly ignores the information regularization for these student-problem pairs. 
Compared to InfoGAN~\cite{chen2016infogan} showing consistent variations in an unsupervised hand-written digit generation setting, we hypothesize that capturing variations in generated code across multiple problems on different topics is harder.  
Since different students attempt different problems, InfoOIRT learns a student-specific $\hat{h}_{\text{disc}}$ distribution where the effect of each dimension changes depending on the attempted problems.

%
%

\subsubsection{Continuous Latent Knowledge Factors}

We manipulate the learned continuous latent factor $\hat{h}_{\text{cont}}$ (with a range between $-3.5$ to $3.5$) for different ranges and investigate the resulting change in generated code. We do not observe any change for small variations ($-2$ to $2$), showing the robustness of the InfoOIRT, balancing generation accuracy and interpretability. For larger variations ($-5$ to $5$), we see new codes generated with variation in either style, correctness, bug type, or structure, for some students in some problems. For bugs, these changes in the continuous latent variable overlap with flipping certain binary classes among the discrete factors. This observation reflects the nature of code being more discrete rather than continuous. For extreme variations ($-10$ to $10$) that extrapolate $\hat{h}_{\text{cont}}$ beyond learned values, we observe that InfoOIRT still generates coherent student code but interestingly, to a \emph{different problem}. We see that student code for easier problems with conditionals changes to code for harder problems with looping, and vice versa, as shown in Table~\ref{tab:qualitative_results}. This observation suggests that $\hat{h}_{\text{cont}}$ could be discretizing the student's knowledge space of unique code constructs across problem types.
 
%
%
%
%

\subsection{Potential Use Cases in CS Education}
\label{sec:use_case_cs_ed}
Predicting free-form student responses to open-ended problems provides educators a deeper insight into a student's reasoning process~\cite{anderson,brown,feldman,smith} through their knowledge states. Doing so can potentially shed light on the typical errors among students before even assigning them, which enables educators to anticipate and prepare corresponding feedback. With informative and interpretable latent states, it can be easier for intelligent tutoring systems to support educators by summarizing common bugs and coding styles among students. Information on latent factors that indicate student bugs can potentially be used to quantize the effectiveness of additional instruction on different topics on helping students correct errors, which may help educators plan their activities. Moreover, for latent codes that we uncover to associate with specific student bugs, we can explore using them to provide progressive edit suggestions, by gradually changing the continuous latent variables to generate code between a student's original buggy code and correct code, which can be both informative and relatable to the student. 

%
%
%
%
%
%

\section{Conclusions and Future Work}
We presented a first step towards interpreting latent student knowledge states in models of open-ended responses in programming education. We proposed InfoOIRT, an open-ended IRT model that accurately predicts student-written code, validated on the real-world CSEDM dataset, along with interpretable latent student knowledge states.
Through qualitative analysis, we presented examples of latent student knowledge states capturing salient syntactic and semantic features including style, mastery of programming skills, and code structure, demonstrating the potential of InfoOIRT in CS education.
InfoOIRT should be considered exploratory with limitations and several avenues for future work. First, we can explore adapting InfoOIRT to knowledge tracing and impose constraints on the consistency of these states over time, using ideas from cognitive modeling~\cite{shi2023kc}. 
Second, we can reduce potential biases toward underrepresented students by minimizing the mutual information between demographic variables and student-written code. 
Third, we can explore applying InfoOIRT to other domains including language learning~\cite{jake-language-learning}, and mathematics~\cite{scarlatos-lan-2023-tree}.

%
%
%
%
%
%

\section{Acknowledgements}
We thank Alexander Scarlatos and the anonymous reviewers for their helpful comments. We thank the NSF for partially supporting this work under grants DUE-2215193 and IIS-2237676.

%
\bibliographystyle{abbrv}
\bibliography{sigproc}  
%

\clearpage
\appendix

%
%
%
%
%

\section{Experiments}
\label{sec:supp_experiments}

\subsection{Dataset}
We ground our analysis on the real-world programming education dataset from the $2$nd CSEDM Data Challenge, which we referred to as the CSEDM dataset.\footnote{\url{https://sites.google.com/ncsu.edu/csedm-dc-2021}} This dataset contains $46,825$ student code submissions from $246$ college students on $50$ open-ended Java programming problems collected over an entire semester. We analyze the first submission to each problem and ignore later attempts since this setting captures a student's overall mastery of programming concepts while later attempts also capture debugging skills that we do not analyze in this work. We preprocess the dataset by removing $15\%$ of code submissions that cannot be converted to an abstract syntax tree (AST) and split it into train-validation-test with $80\%-10\%-10\%$ proportion.

\subsection{Metrics}
Following OKT~\cite{okt}, we quantitatively evaluate generated student code using the popular metric CodeBLEU~\cite{codebleu}, which measures syntactic and semantic similarity between generated and actual student codes. We also report the average test loss of GPT-2 across generated code tokens using the model with the lowest validation loss with a lower test loss being predictive of better student code generation performance. To test whether OIRT simply memorizes the training data, we measure diversity in the generated student codes using the dist-N metric~\cite{distn}, which computes the ratio of unique $N$-grams in the generated codes over all N-grams, with $N=1,2,3$.

\subsection{Implementation Details}
 
For OIRT training, we follow the setup in~\cite{okt} and use a batch size of 8, an AdamW optimizer with a learning rate of $1\cdot 10^{-5}$ with linear learning rate scheduler with warmup for GPT-2 model parameters, and the Adam optimizer with a learning rate of $1\cdot 10^{-3}$ for the alignment function and student specific $\bar{h}$ knowledge states. We finetune OIRT for $50$ epochs which takes around $4$ hours on a single NVIDIA A100 GPU, and chose the model with the lowest validation loss.

For InfoOIRT, we chose to model the latent factors in student knowledge states $\hat{h}$, with $1$ continuous dimension and $10$ discrete dimensions, each having two classes and thereby encouraged to act as binary switches of programming knowledge mastery states/syntactic styles. 
We learn a student-specific $2$-dimensional static knowledge state $\bar{h}$. For a fair comparison with OIRT, we learn a student-specific $23$ dimensional $\bar{h}$ in OIRT and use the same hyperparameters in both models.
Since our goal is to analyze the change in code generated with respect to variation in latent factors $\hat{h}$ only, to remove randomness during inference, we use greedy decoding to generate student code in both models.

\end{document}

%% file: student_codes/conciseness_11.tex
\begin{lstlisting}
public int caughtSpeeding(int speed, boolean isBirthday)
{
    if (isBirthday)
        if (speed <= 65)
            return 0;
        else if (speed >= 66 && speed <= 85)
            return 1;
        else
            return 2;
    else
        if (speed <= 60)
            return 0;
        else if (speed >= 61 && speed <= 80)
            return 1;
        else
            return 2;
}\end{lstlisting}

%% file: student_codes/conciseness_12.tex
\begin{lstlisting}
public int caughtSpeeding(int speed, boolean isBirthday)
{
    if(isBirthday)
        speed -= 5;
    if(speed <= 60)
        return 0;
    else if(speed <= 80)
        return 1;
    else
        return 2;
}\end{lstlisting}

%% file: student_codes/conciseness_41.tex
\begin{lstlisting}
public boolean cigarParty(int cigars, boolean isWeekend)
{
    if (isWeekend)
        if (cigars >= 40)
            return true;
        else
            return false;
    else
        if (cigars >= 40 && cigars <= 60)
            return true;
        else
            return false;
}\end{lstlisting}

%% file: student_codes/conciseness_42.tex
\begin{lstlisting}
public boolean cigarParty(int cigars, boolean isWeekend)
{
    if(isWeekend)
        return (cigars >= 40);
    return (cigars >= 40 && cigars <= 60);
}\end{lstlisting}

%% file: student_codes/style_11.tex
\begin{lstlisting}
public boolean isEverywhere(int[] nums, int val)
{
    for (int i = 0; i < nums.length; i++) {
        if (nums[i]!= val && nums[i+1]!= val) {
            return false;
        }
    }
    return true;
}\end{lstlisting}

%% file: student_codes/style_12.tex
\begin{lstlisting}
public boolean isEverywhere(int[] nums, int val)
{
    for (int i = 0; i < nums.length; i++)
    {
        if (nums[i]!= val && nums[i+1]!= val)
        {
            return false;
        }
    }
    return true;
}\end{lstlisting}

%% file: student_codes/style_41.tex
\begin{lstlisting}
public int makeChocolate(int small, int big, int goal)
{
    int maxBig = goal/5;
    if (maxBig <= big) {
        goal -= maxBig*5;
    }
    else {
        goal -= big*5;
    }
    if (goal <= small) {
        return goal;
    }
    return -1;
}\end{lstlisting}

%% file: student_codes/style_42.tex
\begin{lstlisting}
public int makeChocolate(int small, int big, int goal)
{
    int maxBig = goal/5;
    if (maxBig <= big)
    {
        goal -= maxBig*5;
    }
    else
    {
        goal -= big*5;
    }
    if (goal <= small)
    {
        return goal;
    }
    return -1;
}\end{lstlisting}

%% file: student_codes/problem_type_11.tex
\begin{lstlisting}
public boolean squirrelPlay(int temp, boolean isSummer)
{
    if(isSummer)
        return (temp >= 60 && temp <= 100);
    return (temp >= 60 && temp <= 90);
}\end{lstlisting}

%% file: student_codes/problem_type_12.tex
\begin{lstlisting}
public boolean xyBalance(String str)
{
    int len = str.length() - 1;
    char ch;
    for(int i = len; i >= 0; i--)
    {
        ch = str.charAt(i);
        if(ch == 'x')
            return false;
        else if(ch == 'y')
            return true;
    }
    return true;   
}\end{lstlisting}